\pdfoutput=1

\documentclass[11pt]{article}

\usepackage[final]{acl}

\usepackage{times}
\usepackage{latexsym}
\usepackage{amsmath}
\usepackage{amssymb}
\usepackage{enumitem}

\usepackage[T1]{fontenc}

\usepackage[utf8]{inputenc}

\usepackage{microtype}

\usepackage{inconsolata}

\usepackage{graphicx}
\usepackage{todonotes}
\usetikzlibrary{shadows}

\usepackage{multirow}
\usepackage{tabularx}
\usepackage{ragged2e}
\usepackage{booktabs}
\usepackage{makecell}

\title{Supporting Online Discussions: Integrating AI Into the adhocracy+ Participation Platform To Enhance Deliberation}

\author{Maike Behrendt \\
	Heinrich Heine University\\
	\small{\texttt{maike.behrendt@hhu.de}} \\\And
	Stefan Sylvius Wagner \\
	Heinrich Heine University\\    
	\small{\texttt{stefan.wagner@hhu.de}} \\\\\And
    Mira Warne \\
    Heinrich Heine University \\
    \small{\texttt{mira.warne@hhu.de}} \\\AND
    Jana Leonie Peters \\
    Heinrich Heine University \\
    \small{\texttt{jana.leonie.peters@hhu.de}}\\\And
    Marc Ziegele \\
    Heinrich Heine University \\
    \small{\texttt{marc.ziegele@hhu.de}}\\\And
	Stefan Harmeling \\
	Technical University Dortmund\\
	\small{\texttt{stefan.harmeling@tu-dortmund.de}}}

\begin{document}
\maketitle
\begin{abstract}
Online spaces provide individuals with the opportunity to engage in discussions on important topics and make collective decisions, regardless of their geographic location or time zone. However, without adequate support and careful design, such discussions often suffer from a lack of structure and civility in the exchange of opinions. Artificial intelligence (AI) offers a promising avenue for helping both participants and organizers in managing large-scale online participation processes. This paper introduces an extension of adhocracy+, a large-scale open-source participation platform.  Our extension features two AI-supported debate modules designed to improve discussion quality and foster participant interaction.
In a large-scale user study we examined the effects and usability of both modules. We report our findings in this paper. The extended platform is available at \url{https://github.com/mabehrendt/discuss2.0}.
\end{abstract}


\section{Introduction}


Online discussions and participation platforms enable people to engage in socially relevant issues. However, written exchanges in online spaces are frequently marked by a lack of structure, often leading to \emph{information overload}, making it difficult for both participants and providers to process large volumes of contributions~\citep{arana2021citizen}. According to \citet{10.1145/3593743.3593771}, other key issues include polarization, incivility, toxic behavior, superficial content, and insufficient collaboration among participants.
To address these challenges, the concept of deliberation proves particularly valuable.
Deliberation is defined as the respectful and argumentative exchange of opinions aimed at making a decision. It encompasses three core dimensions: \emph{rationality}, referring to the argumentative exchange of opinions; \emph{civility}, which entails politeness and respect; and \emph{reciprocity}, characterized by responsiveness and active listening~\citep{bachtiger2009measuring,esau2021different,10.1007/978-3-642-15158-3_3}.

AI presents a promising opportunity to enhance deliberation, supporting both participants and organizers in creating a more structured, respectful, and engaging environment for meaningful exchange of opinions. In this work, we propose two AI-based solutions to improve online discussions, implemented for adhocracy+, an open-source participation platform.

\paragraph{Our contributions:}
\begin{enumerate}
    \item \textbf{Comment Recommendation Module:} To encourage user interaction and expose participants to opposing viewpoints, we developed a comment recommendation module based on a stance detection model.
    \item \textbf{Deliberative Quality Module:} To enhance user engagement and improve the quality of contributed comments, we implemented a debate module that automatically detects and highlights the most deliberative comments.
    \item \textbf{Application and Evaluation:} To examine the effects of the proposed modules, we conducted a large-scale panel study (N~=~1,356). The results of the user study are presented in detail in the following sections.
\end{enumerate}




\begin{figure*}[t!]
  \centering
  \begin{tikzpicture}    
    \node[draw, thick, color=black, fill=white, anchor=south west,inner sep=3mm, drop shadow] (image1) at (0,0)
        {\includegraphics[width=0.45\textwidth]{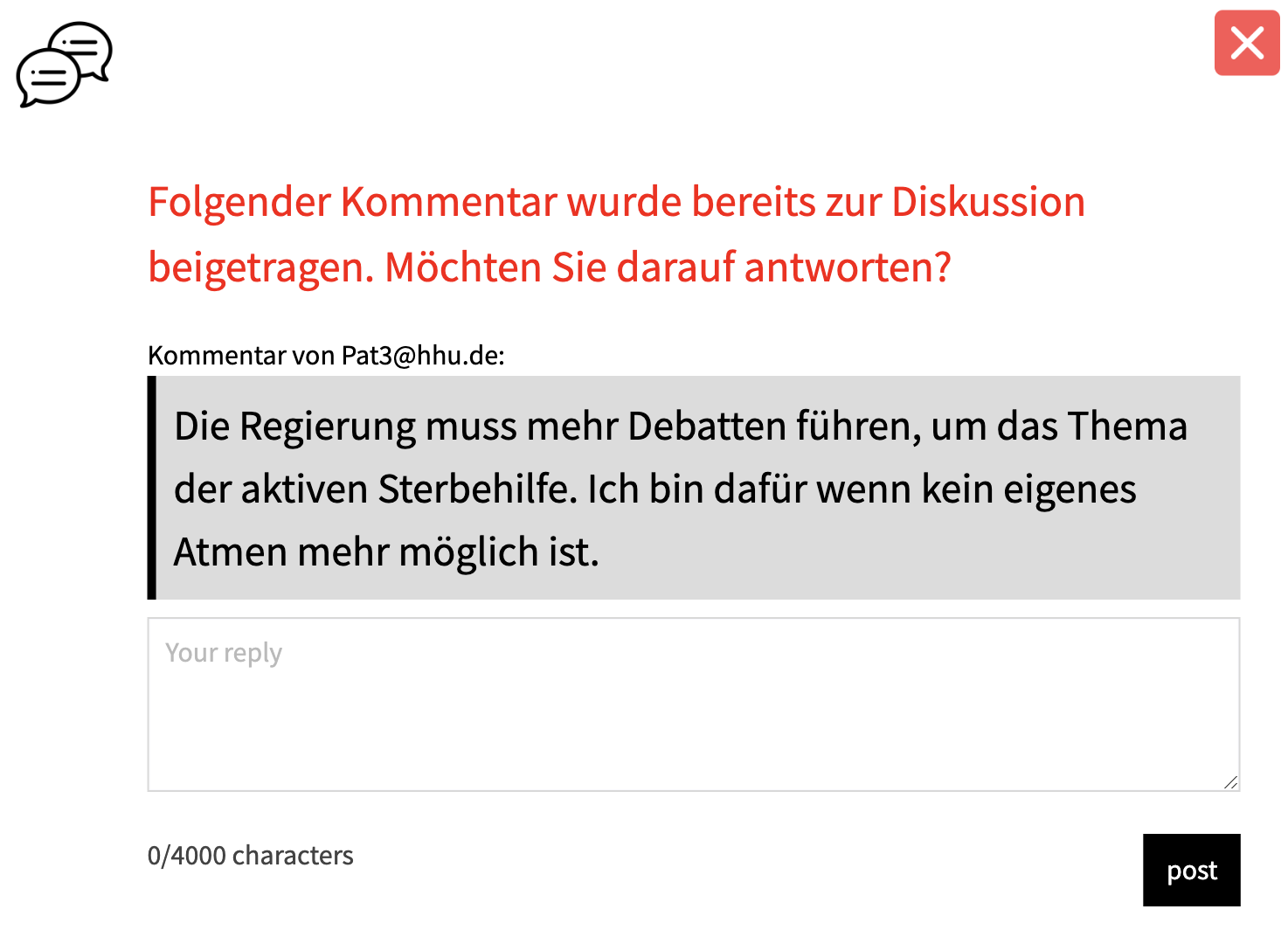}};
    \hspace{2.5mm}
    \node[draw, thick, color=black, fill=white, anchor=south west,inner sep=2mm, drop shadow] (image2) at (image1.south east)
        {\includegraphics[width=0.44\textwidth]{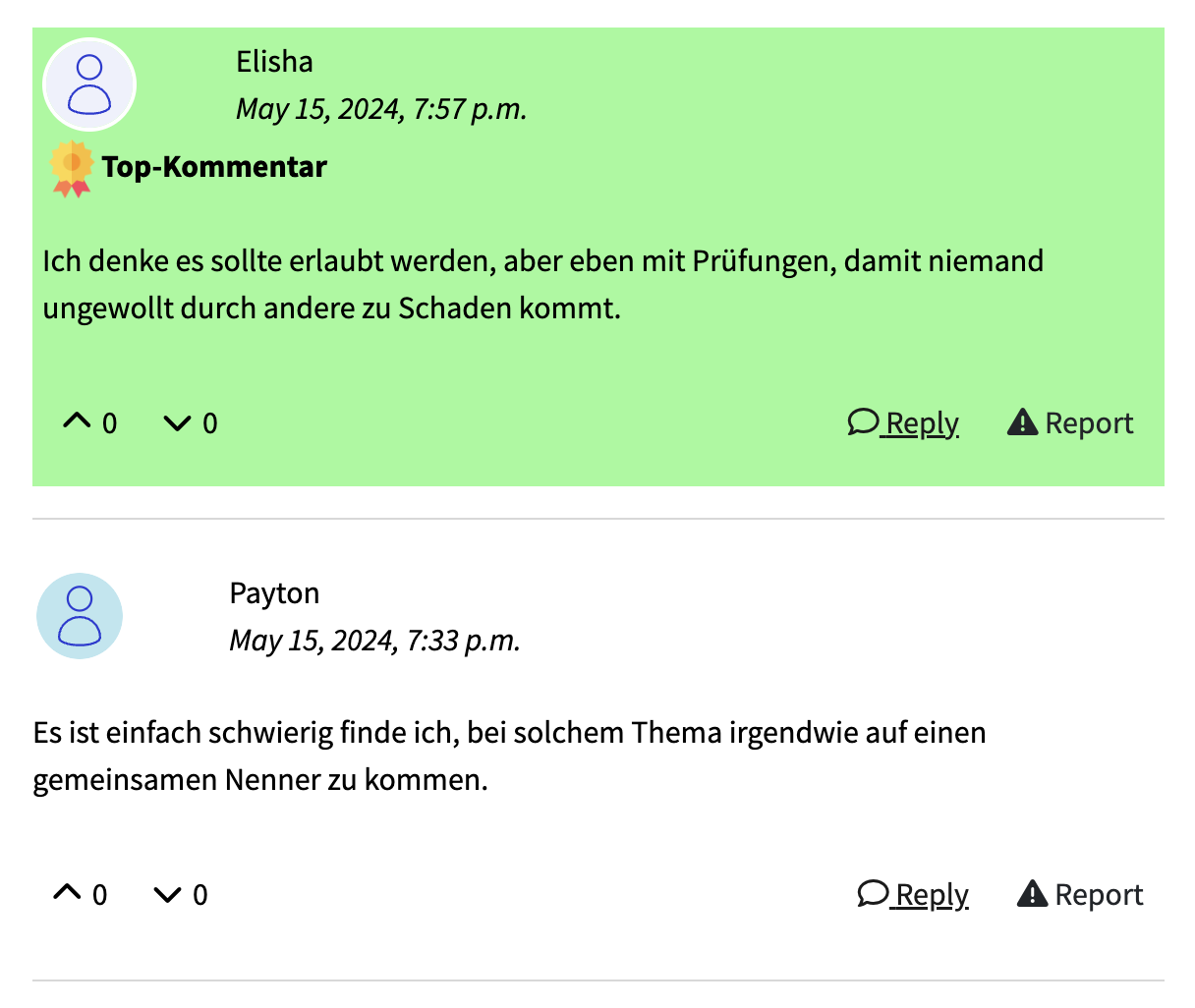}};
  \end{tikzpicture}
  \caption{We propose two AI tools that we integrate into adhocracy+. \textbf{(Left) Comment Recommendation Module:} Participants are confronted with a comment that contradicts their own opinion and are asked if they want to respond. The AI tool determines the stance of the comments, which is used to propose opposing comments. Translation: The following comment has already been added to the discussion. Do you want to reply to it? \textbf{(Right) Deliberative Quality Module:} We predict a deliberative quality score (AQuA score) for each comment. Comments with a high AQuA score are sorted to the top of the discussion and highlighted in bright green and marked as "top comment".}
  \label{fig:screenshots}
\end{figure*}

\section{Related Work}


Previous efforts to integrate AI into discussion platforms have often focused on structuring and summarizing discussions.
The CONSUL\footnote{\url{https://consulproject.nl/en/}} citizen participation tool enables citizens to propose ideas to local politicians on improving their city. These proposals can be supported and discussed by other participants on the platform. To address the issue of \emph{information overload}, \citet{arana2021citizen} improved the platform with several natural language processing (NLP) methods, including tools to summarize existing proposals, automatically categorize them and recommend proposals to participants according to their interests.

In the KOSMO project, an AI-supported moderation dashboard was developed for the adhocracy+ platform to assist moderators during citizen participation processes\footnote{\url{https://github.com/liqd/a4-kosmo}}. Two models were trained to identify uncivil and engaging comments~\citep{risch-etal-2021-overview}, which are flagged for moderators, allowing them to decide on appropriate actions, such as blocking uncivil comments.

The BCause platform, created by \citet{10.1145/3593743.3593771}, supports discussions with an automatic text summarization tool and an argument recommendation system. This system suggests arguments from scientific literature based on the user's stance on the discussed topic. Other examples of open-source discussion tools that incorporate AI features include Discourse\footnote{\url{https://meta.discourse.org/}} and Polis~\citep{small2021polis} from the Computational Democracy Project.

Another notable example is CommunityPulse~\citep{10.1145/3461778.3462132}, a platform equipped with tools for text analysis and visualization to help civic leaders to make sense of community input. It includes a sentiment analysis of contributions and topic modeling to automatically extract discussed topics.

Beyond civic tech, there is a broader body of research focused on using AI to support discussion in the context of collaborative learning (see, e.g., \citet{KONG2025100973}). 

Similar to our approach, \citet{10.1145/3613904.3642530} also aim to enhance deliberative quality on online discussion platforms. They employ large language models (LLMs) to generate reflective nudges designed to promote users’ self-reflection, thereby fostering more thoughtful and deliberative contributions. In our work, we focus on directly enhancing the deliberative quality of discussions by improving their reciprocity and rationality. To achieve this, we introduce two new modules for the adhocracy+ platform: (i)~the \emph{Comment Recommendation Module} that suggests comments based on whether participants are \emph{in favor} or \emph{against} the discussed issue, encouraging participants to engage with opposing viewpoints, and (ii)~the \emph{Deliberative Quality Module} that automatically identifies and highlights the most deliberative user comments, motivating participants to contribute further high-quality comments.




\begin{figure*}[t!]
  \centering
  \begin{tikzpicture}    
    \node[draw, thick, color=white, fill=white, anchor=south west,inner sep=0mm] (image1) at (0,0)
        {\includegraphics[width=0.33\textwidth]{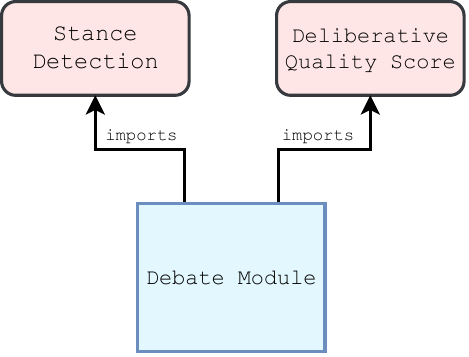}};
    \hspace{2mm}
    \node[draw, thick, color=white, fill=white, anchor=south west,inner sep=2mm] (image2) at (image1.south east)
        {\includegraphics[width=0.65\textwidth]{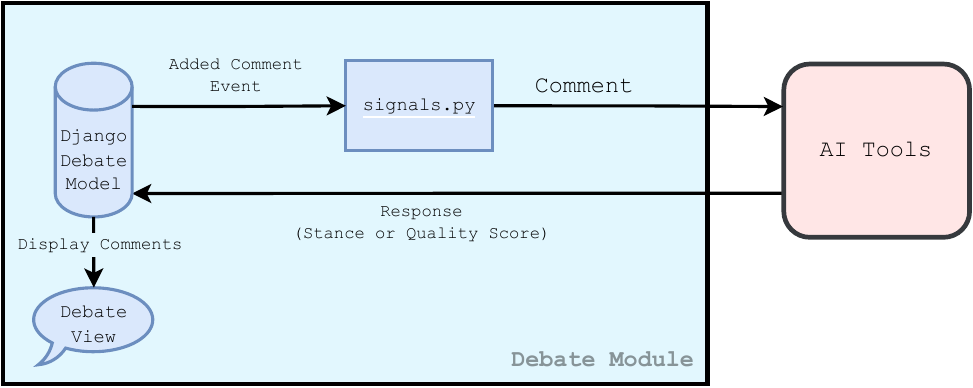}};
  \end{tikzpicture}
\caption{Overview of the architecture to extend adhocracy+ with our AI tools. \textbf{(Left)} The \emph{debate module} imports both the \emph{stance detection} and \emph{deliberative quality} AI's as Python modules. \textbf{(Right)} The Django database model sends out an event when a new comment is added to the database. The event is handled in signals.py where the new comment is passed either to the stance detection or deliberative quality model. These send a response (either a stance or quality score) back to the database where the corresponding response is stored.}
  \label{fig:architecture}
\end{figure*}

\section{Features}
In the following, we will discuss the features of the two implemented modules from both a technical and a user perspective.
\subsection{Enhancing Reciprocity with the Comment Recommendation Module}
As previously mentioned, large-scale online discussions often involve a high volume of postings, including redundant, toxic, or uncivil content. Simultaneously, these discussions frequently lack structure, leading participants to experience information overload~\citep{arana2021citizen}. Under this condition, participants struggle to follow the discussion and engage with others, which results in a lack of reciprocity within the conversation~\citep{lago2019citizen}. Another consequence of information overload is dysfunctional argumentation~\citep{klein2011harvest}. This, in turn, fosters the formation of small groups of participants who share similar opinions and avoid interacting with those holding opposing views~\citep{klein2015critical}.

To mitigate information overload, enhance reciprocity among participants, and improve the overall quality of discussions, we developed a \emph{Comment Recommendation Module} integrated into the adhocracy+ debate module. This module suggests comments to the participants that reflect a point of view opposite to their own. For instance, if a participant holds an \emph{against} stance on the debate question, the module will recommend a comment from another participant with an \emph{in favor} stance.

\paragraph{Stance Detection.} To detect the stance of a comment, we use an uncased German BERT Base model~\citep{chan-etal-2020-germans}\footnote{\url{https://huggingface.co/dbmdz/bert-base-german-uncased}} fine-tuned on the X-Stance dataset~\citep{vamvas2020xstance}. This dataset includes 48.6k German comments on 150 political questions, answered by political office candidates in Switzerland. Since the adhocracy+ platform is specifically designed for discussion and decision making on politically relevant issues, the dataset fits our purpose very well. The fine-tuned model operates as a binary classifier, outputting either \emph{in favor} or \emph{against} based on a given debate question and a specific comment. 

The complexity and diversity of political and social issues make it challenging to obtain sufficient labeled data for stance detection. To address this, we follow the approach of~\citet{wagner2025the}, leveraging synthetic data generated by LLMs. We employ Mistral-7B~\citep{jiang2023mistral} to generate comments reflecting an \emph{in favor} or \emph{against} stance. These synthetic comments are then used to further fine-tune the stance detection model. For existing comments, the synthetic data helps identify real comments that are most challenging for the model to classify. These comments can be manually labeled to further improve the model's performance. For additional details, see~\citet{wagner2025the}.

\paragraph{Comment Recommendation User Experience.} The main purpose of the \emph{Comment Recommendation Module} is to present a comment to the user that opposes their own position on the debate question. Therefore, the stance of every comment, posted in the discussion is predicted and stored into the database. When a user logs into the platform for the first time, they are prompted to indicate their stance on the debate question.

The user's position, which can be either \emph{in favor} or \emph{against}, is stored in the database. This information is then used to determine suitable comments for recommendation. The system retrieves comments from the database that oppose the user's stance. If multiple opposing comments are available, one is randomly selected from the list. If there is no suitable comment available, a message is displayed to the user, indicating that no comment can be suggested at that time.

The selected comment is displayed to the user in a popup window  (see Figure~\ref{fig:screenshots}, left), where they are given the option to reply. Once the user responds, the popup dialog closes, and the screen automatically scrolls to the suggested comment within the discussion. Additionally, users can reopen the suggestion popup by clicking a designated icon. When reopened, a new opposing comment (if available) is proposed for the user to reply to.

\subsection{Enhancing Debate Quality and Engagement With the Deliberative Quality Module}

In addition to disorganized content and dysfunctional argumentation (which diminishes reciprocity, see the previous section), online discussions face other challenges, including low-quality contributions~\citep{klein2011harvest}. Addressing this issue is crucial for fostering meaningful and productive conversations. In an observational study, \citet{10.1145/3484245} found that manually highlighting high-quality comments in the comment section of the New York Times (referred to as the \emph{New York Times Picks}) increases the overall discussion quality and the user engagement. The authors suggest that highlighting well-written comments is beneficial to the quality of new comments as the picked comments constitute a social feedback mechanism~\citep{10.1145/3484245}. 

We build on these findings and develop the \emph{Deliberative Quality Module} which aims to promote high quality comments by automatically highlighting them. It remains to be investigated whether the human component, i.e., the selection by a New York Times editor, has a significant impact on the participants' perceptions, or whether simply highlighting the comments has the same effect.
To measure the deliberativeness of a user comment, we calculate the AQuA score~\citep{behrendt-etal-2024-aqua} for each comment and define a threshold for high quality.


\paragraph{AQuA Score.} The AQuA score, proposed by~\citet{behrendt-etal-2024-aqua}, is a weighted sum of the predictions of individual BERT-based adapter models $f_{\theta_k}$~\citep{pfeiffer-etal-2020-adapterhub}, fine-tuned for 20 different deliberative quality indicators. These include, i.a., \emph{justification}, \emph{proposing solutions}, \emph{referencing other users} and, as an indicator for low quality, the use of incivility markers, such as \emph{sarcasm}. Each adapter prediction is weighted with a number $w_k\in\mathbb{R}$ that is estimated from data. Some of the weights are positive, indicating a positive correlation between the respective indicator and the overall quality of the comment, and some are negative, indicating a negative correlation. The total score for a comment $c$ is calculated as
\begin{equation}
    s_{\text{AQuA}}(c) = \sum^{20}_{k=1}w_k f_{\theta_k}(c).
\end{equation}

AQuA scores are normalized to the range between 0 and 5. 
Note that the individual predictions of AQuA are trained on expert evaluations, which are combined with weights estimated from non-expert assessments, for details see \citet{behrendt-etal-2024-aqua}.
In the \emph{Deliberative Quality Module}, AQuA scores allow us to identify high quality comments. 

\paragraph{Deliberative Quality User Experience.}
The three comments with the highest predicted AQuA scores, which exceed a specified threshold, are automatically identified as top comments. They are prominently displayed above the other comments and highlighted in light green color (see Figure~\ref{fig:screenshots} on the right, showing only a single top comment). The other comments are displayed below the top comments in chronological order.
The exact threshold depends on the discussion and can be set as a hyperparameter.

\section{Implementation Details}
Adhocracy+ is built on the Django framework\footnote{\url{https://www.djangoproject.com/}} and provides a wide range of functionalities and modules to facilitate large-scale online discussions. The platform's debate module features a forum-like structure where a discussion topic is defined and displayed at the top of the page, enabling users to comment on the topic or respond to other participants' comments. Additional details about the platform's features are available on the adhocracy+ website\footnote{\url{https://adhocracy.plus/info/features/}}.
We extend adhocracy+ by importing the AI tools into the debate module, as shown in Figure~\ref{fig:architecture} (left). A more detailed view is shown in Figure~\ref{fig:architecture} on the right. When a new comment is added by a user, the Django debate model fires an event, which is handled in the \texttt{signals.py} file. Here, we import the AI tools to pass the comments to the stance detection or the deliberative quality model. The AI tools then return a response (either a stance or deliberative quality score), which is stored back to the database for the corresponding comment. This stored response is then presented by the corresponding module as shown in Figure~\ref{fig:screenshots}. For the purposes of this study, they were implemented as distinct debate modules within the adhocracy+ platform in order to enable the separate evaluation of their respective effects. Their integration into a unified module remains a plausible direction for future development. Overall, this architecture is flexible: In our experiments, we ran the AI tools locally on a Linux server. But the AI tools could also be run as services where communication is handled via Rest API.



\begin{table}[]
    \centering
    \footnotesize
    \begin{tabular}{lcc}
        \toprule
         \textbf{Model} & \textbf{Acc.} & \textbf{F1} \\
         \hline
         BERT Base German Cased & 0.7381 & 0.7426 \\
         \bottomrule
    \end{tabular}
    \caption{The performance on the test set of the X-Stance dataset~\citep{vamvas2020xstance} of the fine-tuned BERT Base German cased model we used for stance prediction.}
    \label{tab:x_stance}
\end{table}
\vspace*{-0.2\baselineskip}
\begin{table}
    \centering
    \small
    \begin{tabular}{clc}
        \toprule
         &\textbf{Deliberative Aspect}&\textbf{MBERT uncased}\\ 
         & \\[\dimexpr-\normalbaselineskip+2pt]
        \hline
         \multirow{7}{*}{\rotatebox[origin=c]{90}{Rationality}\hspace{-3mm}}
         & \\[\dimexpr-\normalbaselineskip+2pt]
         &Relevance&0.37\\
         &Fact&0.56\\
         &Opinion&0.57\\
         &Justification&0.69\\
         &Solution Proposals&0.79\\
         &Additional Knowledge&0.78\\
         &Question&0.87\\
         & \\[\dimexpr-\normalbaselineskip+2pt]
         \hline
         \multirow{5}{*}{\rotatebox[origin=c]{90}{Reciprocity}\hspace{-3mm}}
         & \\[\dimexpr-\normalbaselineskip+2pt]
         &Referencing Users&0.88\\
         &Referencing Medium&0.93\\
         &Referencing Contents&0.81\\
         &Referencing Personal& 0.92\\
         &Referencing Format&0.96\\
         & \\[\dimexpr-\normalbaselineskip+2pt]
         \hline
         \multirow{7}{*}{\rotatebox[origin=c]{90}{Civility}\hspace{-3mm}}
         & \\[\dimexpr-\normalbaselineskip+2pt]
         &Polite form of Address&0.97\\
         &Respect&0.9\\
         &Screaming&0.81\\
         &Vulgar&0.74\\
         &Insults&0.87\\
         &Sarcasm&0.48\\
         &Discrimination&0.88\\
        \hline
        & \\[\dimexpr-\normalbaselineskip+2pt]
         &Storytelling&0.85\\
         & \\[\dimexpr-\normalbaselineskip+2pt]
         \hline 
        & \\[\dimexpr-\normalbaselineskip+2pt]
        &\O \, Total Average (F1-Score) &0.7815\\
        \bottomrule
    \end{tabular}
    \caption{We show the weighted average F1 score for the 20 different deliberative aspects the AQuA score adapter models are trained on.}
    \label{tab:aquaf1}
\end{table}

\section{Evaluation}
In the following, we analyze the effectiveness of our two proposed modules. We start by evaluating both models on existing datasets and measure how well they perform in terms of accuracy and F1 score. Furthermore, we conducted a large-scale user study to evaluate participants' satisfaction when using the modules in a real online discussion as well as to gauge the effects of the modules on other perceptions and behaviors of the participants.

\subsection{Model Performance}
\paragraph{Comment Recommendation Module}

Table~\ref{tab:x_stance} displays the performance of the German BERT Base uncased model, which was fine-tuned on the X-Stance dataset \cite{vamvas2020xstance}. The model reaches an accuracy of $73.81$ and an F1 score of $74.26$ on the test dataset.



\paragraph{Deliberative Quality Module}

A multilingual BERT base uncased model\footnote{\url{https://huggingface.co/google-bert/bert-base-multilingual-cased}} serves as the basis for the trained adapter models that build the AQuA score~\citep{behrendt-etal-2024-aqua}. Table~\ref{tab:aquaf1} lists the weighted average F1 scores on the test dataset for each of the 20 trained adapter models on deliberative aspects.

\begin{table*}[h]
    \small
    \centering
    \renewcommand{\arraystretch}{1.2}
    \setlength{\tabcolsep}{5pt}
    \begin{tabularx}{\textwidth}{l X l}
        \toprule
        \textbf{\#} & \textbf{Survey Question} & \textbf{Scale (1-7)} \\
        \midrule
        Q1 & On the platform, discussion contributions were suggested to me, to which I could reply. & 1 = strongly disagree, 7 = strongly agree\\
        Q2 & On the platform, contributions were marked as top comments. &1 = strongly disagree, 7 = strongly agree\\
        Q3 & To what extent did you feel that this process was supported by artificial intelligence? & 1 = most certainly not, 7 = most certainly yes\\
        $\text{Q4}^*$ & I enjoyed using discuss20. & 1 = strongly disagree, 7 = strongly agree\\
        $\text{Q5}^*$& The functions of the discuss20 platform threatened my freedom to choose what I wanted. & 1 = strongly disagree, 7 = strongly agree\\
        Q6 & All in all, I was satisfied with the discussion. &1 = strongly disagree, 7 = strongly agree\\
        $\text{Q7}^*$ & The contributions contained arguments and justifications. & 1 = strongly disagree, 7 = strongly agree \\
        $\text{Q8}^*$ & The participants responded to the contributions of others. & 1 = strongly disagree, 7 = strongly agree \\
        $\text{Q9}^*$ & The contributions were discriminating. & 1 = strongly disagree, 7 = strongly agree\\
        $\text{Q10}^*$ & There was a wide range of opinions in the discussion. & 1 = strongly disagree, 7 = strongly agree\\
        \bottomrule
    \end{tabularx}
    \caption{Excerpt from our user study survey questions. Questions that are marked with an asterisk are example questions that are part of a larger index.}
    \label{tab:survey}
\end{table*}

\subsection{User Study}
\subsubsection{Methodology}
To investigate the effects of both AI modules, we conducted a field experiment as part of a three-wave panel survey in July 2024. Participants were recruited from the German population through Bilendi, an online access panel provider and market research company. The final sample consisted of N = 1,356 participants with a mean age of 52 years (47\% female; 58\% with at least a high school diploma). 

Participants joined a simulated citizens’ assembly with a 10-day online discussion phase on the extended adhocracy+ platform (internally referred to as \textit{discuss20}). They engaged in small-group discussions on two selected political topics: (1)~whether active euthanasia should be legally permitted in Germany, and (2)~whether the sale of alcoholic beverages should be more restricted in Germany. These topics were identified in a preliminary survey as the most engaging from a broader selection of issues.

The experimental design consisted of five conditions for each of the two discussed topics, aimed at testing the effects of the AI modules. These included: discussions supported by the \emph{Comment Recommendation Module}, which either (i) recommended comments that contradicted the participant’s opinion or (ii) recommended random comments. Discussions supported by the \emph{Deliberative Quality Module}, which either (iii) highlighted three comments with the highest deliberative quality scores as "top comments" or (iv) highlighted three randomly selected comments as "top comments" and (v) discussions without AI support, serving as the control group. Participants and experimental conditions were randomly assigned, resulting in ten distinct experimental groups. Randomization checks showed no significant differences between the groups in terms of age, gender, education, or political interest.

\begin{table*}[t]
    \footnotesize
    \centering
    \begin{tabularx}{\textwidth}{Xccccccc}
    \toprule
     & \multicolumn{2}{c}{\thead{\textbf{AI}\\\textbf{CR Module}\\(n = 289)}} & \multicolumn{2}{c}{\thead{\textbf{Random}\\\textbf{CR Module}\\(n = 276)}} & \multicolumn{2}{c}{\thead{\textbf{Control}\\(n = 262)}} & \multirow{2}{*}{\textbf{F}}\\
      & \textit{M} & \textit{SD} & \textit{M} & \textit{SD} & \textit{M}& \textit{SD} \\
      \midrule
      \textbf{(1) Manipulation effectiveness} \\
      Discussion contributions were suggested to me, to which I could reply & 6.20\textsuperscript{a} & 1.29 & 5.88\textsuperscript{a} & 1.54 & 3.83\textsuperscript{b} & 2.22 & 116.18*** \\
      To what extent did you feel that the discussion was supported by artificial intelligence? & 4.33\textsuperscript{a} & 1.65 & 4.18\textsuperscript{a} & 1.62 & 3.80\textsuperscript{b} & 1.71 & 7.47***\\
      \textbf{(2) Quantity of participation}\\
      Average number of comments per user & 12.71\textsuperscript{a} & 12.85 & 12.98\textsuperscript{a} & 11.85 & 9.16\textsuperscript{b} & 10.58&9.82***\\
      \textbf{(3) Platform evaluation}\\
      Overall satisfaction with the platform & 6.08 & 1.13 & 6.11 & 1.06 & 6.15 & 0.97 & 0.34\\
      Experience of threats to freedom of choice & 1.44&0.89&1.51&0.96&1.37&0.80&1.61 \\
      \textbf{(4) Discussion evaluation}\\
      Satisfaction with the discussion & 5.98\textsuperscript{a} & 1.16 & 5.89\textsuperscript{ab}&1.33&5.63\textsuperscript{b}&1.49&4.60*\\
      Perception of diversity &6.03\textsuperscript{a}&0.94&5.89\textsuperscript{ab}&1.01&5.74\textsuperscript{b}&1.05&5.50**\\
      Perception of reciprocity & 5.64\textsuperscript{a}&1.03&5.41\textsuperscript{a}&1.13&4.97\textsuperscript{b}&1.35&21.00***\\
      \bottomrule
      \multicolumn{8}{l}{n = 827, One-Way ANOVA (Post-Hoc-Test: Bonferroni/Games-Howell), *p<0.05, ** p<0.01, *** p<0.001.}\\
      \multicolumn{8}{l}{Note: Groups with different code letters (a, b) differ significantly at the 5\% level.}\\
    \end{tabularx}
    \caption{Results of One-Way Analyses of Variance (ANOVAs) for the Comment Recommendation (CR) Module.}
    \label{tab:crmodule}
\end{table*}

During and after the discussions, the participants completed standardized online questionnaires to evaluate their experiences on the platform. To explore the effects of the AI modules, this user study focuses on four aspects: 
\begin{enumerate}
    \item \textbf{Manipulation effectiveness} - the extent to which participants recognized and responded to the implemented AI features.
    \item \textbf{Quantitative participation} - the extent of the engagement of the participants in the discussions.
    \item \textbf{Platform evaluation} - users' perceptions of the platforms usability and functions.
    \item \textbf{Discussion evaluation} - participants' assessments of the discussion quality, including satisfaction and deliberative characteristics.
\end{enumerate}

An excerpt of the corresponding survey questions is listed in Table~\ref{tab:survey}. The effectiveness of the manipulations was measured through participants’ recognition of the module-specific functions (see Q1 and Q2) and their assessment of the AI-support (see Q3). Evaluation of the platform included overall satisfaction with the platform (mean index of 5 items, see, e.g., $\text{Q4}^*$, Cronbach’s alpha = .86) as well as evaluation of the functions against the backdrop of freedom of choice (perceived autonomy, mean index of 4 items, see, e.g., $\text{Q5}^*$, Cronbachs Alpha = .85). Lastly, evaluation of the discussions included overall satisfaction with the discussion (see Q6) and the perceived deliberative quality, evaluated across four dimensions, namely the perception of the rationality (mean index of 4 items, see, e.g., $\text{Q7}^*$, Cronbach’s alpha = .85), reciprocity (mean index of 3 items, see, e.g., $\text{Q8}^*$, Cronbach's alpha = .89), civility (mean index of 4 items, e.g., $\text{Q9}^*$, Cronbach's alpha .78) and diversity of the discussions (mean index of 4 items, e.g., $\text{Q10}^*$, Cronbach's alpha .88).

As the \emph{Comment Recommendation Module} aims to expose users to diverse viewpoints, it fosters diversity and reciprocity by encouraging interaction with opposing opinions. In contrast, the \emph{Deliberative Quality Module} promotes civility and rationality by highlighting comments that exemplify high deliberative quality, thereby setting a constructive standard for discussion. Consequently, the analysis focuses on diversity and reciprocity for the \emph{Comment Recommendation Module} and on civility and rationality for the \emph{Deliberative Quality Module}, as these dimensions best capture the intended effects of each intervention.


\begin{table*}[t]
    \footnotesize
    \centering
    \begin{tabularx}{\textwidth}{Xccccccc}
    \toprule
     & \multicolumn{2}{c}{\thead{\textbf{AI}\\\textbf{DQ Module}\\(n = 289)}} & \multicolumn{2}{c}{\thead{\textbf{Random}\\\textbf{DQ Module}\\(n = 276)}} & \multicolumn{2}{c}{\thead{\textbf{Control}\\(n = 262)}} & \multirow{2}{*}{\textbf{F}}\\
      & \textit{M} & \textit{SD} & \textit{M} & \textit{SD} & \textit{M}& \textit{SD} \\
      \midrule
      \textbf{(1) Manipulation effectiveness} \\
      On the platform, contributions were marked as top comments & 5.81\textsuperscript{a} & 1.76 & 5.90\textsuperscript{a} & 1.61 & 3.05\textsuperscript{b} & 2.02 & 189.17*** \\
      To what extent did you feel that the discussion was supported by artificial intelligence? & 4.20\textsuperscript{a} & 1.80 & 4.50\textsuperscript{a} & 1.61 & 3.80\textsuperscript{b} & 1.71 & 11.22***\\
      \textbf{(2) Quantity of participation}\\
      Average number of comments per user & 9.40 & 10.39 & 9.58 & 9.69 & 9.16 & 10.58&0.12\\
      \textbf{(3) Platform evaluation}\\
      Overall satisfaction with the platform. & 6.16 & 1.02 & 6.06 & 1.13 & 6.15 & 0.97 & 0.74\\
      Experience of threats to freedom of choice & 1.42&0.85&1.39&0.91&1.37&0.80&0.21 \\
      \textbf{(4) Discussion evaluation}\\
      Satisfaction with the discussion & 5.71 & 1.51 & 5.63&1.45&5.63&1.49&0.27\\
      Perception of civility &6.84&0.49&6.83&0.47&6.77&0.66&0.63\\
      Perception of rationality & 5.67&1.04&5.49&1.05&5.51&0.97&2.53\\
      \bottomrule
      \multicolumn{8}{l}{n = 791, One-Way ANOVA (Post-Hoc-Test: Bonferroni/Games-Howell), *p<0.05, ** p<0.01, *** p<0.001}\\
      \multicolumn{8}{l}{Note: Groups with different code letters (a, b) differ significantly at the 5\% level.}\\
    \end{tabularx}
    \caption{Results of One-Way Analyses of Variance (ANOVAs) for the Deliberative Quality (DQ) Module.}
    \label{tab:dqmodule}
\end{table*}

\subsubsection{Results}
\paragraph{Comment Recommendation Module.}
We conducted One-way Analyses of Variance (ANOVAs) to investigate group-specific manipulation effectiveness, quantitative participation, platform evaluation, and discussion evaluation. A summary of the results for the \emph{Comment Recommendation Module} is provided in Table~\ref{tab:crmodule}. We report mean (M) and standard deviation (SD) and F-Values (F). Regarding manipulation effectiveness, participants in the \emph{Comment Recommendation Modules} scored significantly higher on identifying this platform feature and on perceiving AI support compared to the control group. However, the participants’ assessment whether the discussion was supported by AI did not significantly differ between the modules with random and AI-based comment recommendation. 

Regarding participation, participants in the \emph{Comment Recommendation Modules} wrote an average of approximately three to four more comments per user compared to the control group. Again, it was inconsequential whether the recommended comment was suggested randomly or AI-based. Regarding users’ evaluation of the platform, the \emph{Comment Recommendation Modules} did not impair users’ satisfaction with the platform due to the module-specific implemented functions. Another positive finding is that the \emph{Comment Recommendation Modules} did not restrict participants’ feelings of autonomy. 
In contrast, regarding the effects on discussion evaluation, especially participants in the AI-supported \emph{Comment Recommendation Module} reported a significantly higher satisfaction with the discussion and higher perception of the deliberative dimension of diversity compared to the control group. Finally, comment recommendation significantly increased participants’ perception of reciprocity within the discussion compared to the control group. Regardless of an underlying AI-based recommendation, we found that recommending comments had an overall positive effect on individual participation.

\paragraph{Deliberative Quality Module.} Table~\ref{tab:dqmodule} provides an overview of the ANOVA results for the \emph{Deliberative Quality Module}. Regarding manipulation effectiveness, participants in both the AI \emph{Deliberative Quality} and Random \emph{Deliberative Quality Modules} were significantly more likely to recognize platform contributions marked as top comments and to perceive AI support compared to the control group. However, the participants’ assessment of AI support did not significantly differ between the AI \emph{Deliberative Quality} and Random \emph{Deliberative Quality Modules}. 

In terms of participation quantity, the average number of comments per user did not differ significantly between the groups, suggesting that neither the AI \emph{Deliberative Quality} nor the Random \emph{Deliberative Quality Module} led to an increase in users’ commenting activity. Similarly, for platform evaluation, no significant differences were found in users’ overall satisfaction with the platform or their perceptions of autonomy. Participants across all groups reported similarly high satisfaction and did not feel restricted in their freedom to choose actions on the platform. Finally, regarding discussion evaluation, no significant differences were observed between the groups in terms of satisfaction with the discussion, civility, or rationality. While the modules aimed to enhance discussion quality, their implementation did not result in perceptible changes in these specific evaluative dimensions.

In order to compare the actual quality of the discussions across the different groups, content analyses are currently being conducted. Preliminary results suggest that, for the topic of active euthanasia, the quality of discussions was higher in the \emph{Deliberative Quality Module} than in the other modules. Again, however, it appears that it does not seem to make a difference whether the top comments are selected by the AI or at random. 

\section{Conclusion}
In this work we present extensions to the adhocracy+ platform for citizen participation. We implemented two additional modules to support more deliberative online discussions. In the \emph{Comment Recommendation Module} participants are confronted with opposing views to encourage user interaction, hence improving the reciprocity in the discussion.
The \emph{Deliberative Quality Module} aims to improve the quality of contributed comments by automatically highlighting the most deliberative ones. 

In a large-scale user study, we tested the effects of both AI modules. We found that the \emph{Comment Recommendation Module} increased participation on the platform and improved users’ perception of the deliberative quality of the discussions while not diminishing their sense of autonomy. The \emph{Deliberative Quality Module}, in contrast, did not significantly improve users’ perceptions of the platform or the discussions. Still, there are indications that both modules had a positive influence on the discussions, albeit independently of whether AI was involved or not. 

We see great potential in the features we presented to support human actors in conducting large online discussions. Certainly it remains an open task to improve the AI to a level, where people perceive its selection performance as far superior than random selection. The resulting platform is freely available under an open source license and can hopefully be used for political decision-making in the future. 

\paragraph{Future Work.} 
In the future we want to examine how both AI extensions to adhocracy+ can be further improved. This means gathering and annotating additional conversational data to fine-tune and improve both models. To further evaluate effects of both modules on the comment quality within the discussions, content analyses are currently being carried out.

\section{Limitations}
While our extensions to adhocracy+ introduce AI-driven enhancements, we must acknowledge several limitations.

Currently, the platform and both AI modules are only available in German. This limits accessibility for non-German speaking users and limits the potential for wider adoption.

Moreover, the effectiveness of both AI modules highly depends on the quality of their training data. They may struggle with nuanced or complex discussions, and incorrect predictions can potentially frustrate participants.

The effects we observed were predominantly very small, which may be due to the design of our study. In field experiments, numerous noise factors can influence the outcomes we measured - such as the perceived quality of discussions. At the same time, our experiments offer high external validity, as they were conducted in a realistic setting rather than under artificial laboratory conditions.

The partly non-significant differences between the AI, random, and control conditions may also be attributed to the statistical procedures employed. We used post hoc tests that apply strict corrections for multiple testing, which makes it more difficult to detect statistically significant effects. 

However, when conducting planned contrast analyses, some differences between the AI-supported and Random \emph{Comment Recommendation} and \emph{Deliberative Quality Modules} do reach significance, suggesting that the AI-supported modules were perceived more positively by the participants than those working with random content selection. 

Nonetheless, planned contrasts require more specific a priori hypotheses, which could not be formulated within the scope of this exploratory paper. Developing and testing such hypotheses remains a task for future research.





\bibliography{custom}




\end{document}